\documentclass[letterpaper, 10 pt, conference]{ieeeconf}  

\IEEEoverridecommandlockouts                              

\usepackage{amsmath}    
\usepackage{graphicx}   
\usepackage[lofdepth,lotdepth]{subfig}
\usepackage{tabularx}
\usepackage[font=footnotesize,labelfont=footnotesize]{caption}
\usepackage{verbatim}   
\usepackage{color}      
\usepackage{hyperref} 
\usepackage{multirow}
\usepackage{multicol}
\usepackage{graphicx}
\usepackage{rotating}
\usepackage{array}
\usepackage{xspace}
\usepackage{indentfirst}
\usepackage{amssymb}
\usepackage{float}
\usepackage{algorithm}
\usepackage[algo2e]{algorithm2e}
\usepackage{algorithmic}
\usepackage{comment}
\usepackage{sidecap}
\usepackage{booktabs}
\usepackage{breqn}
\usepackage{cite}
\usepackage{mathtools}
\usepackage{graphicx}
\graphicspath{{./figures/}}
\usepackage{caption}
\usepackage{float}
\usepackage{lipsum}
\usepackage{xcolor}
\usepackage{siunitx}
\usepackage{xspace}
\usepackage{bm}
\usepackage{gensymb}
\let\labelindent\relax
\usepackage[inline]{enumitem}
\usepackage{paralist}
\newcommand\norm[1]{\left\lVert#1\right\rVert}
\DeclareMathOperator*{\argmin}{argmin} 

\newcommand{\gobble}[1]{}

\newcounter{tecounter}
\title{\LARGE \bf
Extrinsic Calibration of a 3D-LIDAR and a Camera 
}

\author{Subodh Mishra$^{1}$, Gaurav Pandey$^{2}$ and Srikanth Saripalli$^{1}$
\thanks{$^{1}$with the Department of Mechanical Engineering, Texas A\&M University
        {\tt\small subodh514@tamu.edu}}%
\thanks{$^{2}$with the Ford Motor Company, USA}%
}

\begin{document}

\maketitle
\thispagestyle{empty}
\pagestyle{empty}

\begin{abstract}
This work presents an extrinsic parameter estimation algorithm between a 3D LIDAR and a Projective Camera using a marker-less planar target, by exploiting Planar Surface Point to Plane and Planar Edge Point to back-projected Plane geometric constraints. The proposed method uses the data collected by placing the planar board at different poses in the common field of view of the LIDAR and the Camera. The steps include, detection of the target and the edges of the target in LIDAR and Camera frames, matching the detected planes and lines across both the sensing modalities and finally solving a cost function formed by the aforementioned geometric constraints that link the features detected in both the LIDAR and the Camera using non-linear least squares. We have extensively validated our algorithm using two Basler Cameras, Velodyne VLP-32 and Ouster OS1 LIDARs.
\end{abstract}


\section{Introduction} 

Cameras and LIDARs help robots perceive the environment by providing complementary information. Cameras lack depth information which LIDARs provide and LIDARs lack color, texture and appearance information which cameras provide, hence systems using both these sensing modalities can compensate each others' weaknesses by leveraging respective advantages. A camera can be used for recognizing objects but it doesn't tell how far an object is, such information can be obtained from a LIDAR. 

In addition to perception, cameras and LIDARs are being used together for multi-sensor state estimation, (semantic) mapping and localization. Cameras are better sensors when compared to LIDARs for place recognition and loop closures but are affected by illumination. LIDARs are immune to illumination changes. Simultaneous Localization and Mapping (SLAM) algorithms need both Cameras and LIDARs for robust state estimation. 

Hence, the knowledge of the extrinsic calibration between these sensors is of paramount importance for fusing information from both these sensing modalities. Most perception and state estimation algorithms assume the extrinsic calibration to be known \emph{a-priori}. This calls for methods to estimate the Euclidian 6 Degrees of Freedom (DoF) transformation between a LIDAR center and a Camera center using data generated by them.

\begin{figure}[!ht]
    \centering
    \includegraphics[width=0.475\textwidth]{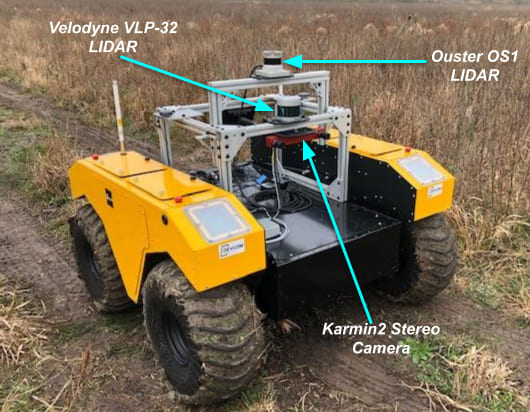}
    \caption{\textbf{Experimental Platform:} Clearpath Robotics Warthog UGV with $a).$ Ouster OS1 LIDAR, $b).$ Velodyne VLP-32 LIDAR and $c).$ Karmin2 Stereo Camera.
    }
    \label{fig:warthogwithsensors}
\end{figure}

\section{Related work}

In literature, the 3D-LIDAR Camera extrinsic calibration problem can be classified into two broad categories \begin {enumerate*} [label=\itshape\alph*\upshape)]
\item Target based approaches and \item Target less approaches
\end {enumerate*}. Target less approaches can be further divided into \begin {enumerate*} [label=\itshape\roman*\upshape)]
\item Scene based approaches and \item Motion based approaches. 
\end {enumerate*}
Although target-less methods are convenient to use, they often rely on a good initial guess, and are also sensitive to the calibration environment. In this paper we focus only on target based 3D-LIDAR Camera calibration because they are computationally light weight and provide good initialization for target less approaches if available. Target based approaches simplify the data association problem. In the works described in \cite{unnikrishnan2005} and \cite{scaramuzza}, the authors have to manually label correspondence between the camera and LIDAR. In this paper the feature association is done automatically. 

Most target based 3D-LIDAR Camera extrinsic calibration methods draw their inspiration from 2D-LIDAR Camera extrinsic calibration techniques. One of the first 2D-LIDAR Camera calibration methods is presented in \cite{Zhang1389752} which relies on observation of a planar checkerboard by both the sensors. The checkerboard gives the pose of the planar surface in the camera frame. The extrinsic parameters between the Camera and LIDAR is determined by solving a geometric constraint formed by projecting the LIDAR points on the checkerboard plane in the camera coordinate system. Observations from several view points need to be taken to ensure that all the DoF are observable. \cite{Naroditsky2011} calibrate a 2D LIDAR Camera system using a geometric constraint that doesn't require the knowledge of the planar target in the camera frame. It rather uses the back-projected plane (formed by a line detected on the image and the camera center in the camera frame) and a LIDAR point lying on the aforementioned line (that defines the back-projected plane) to form a geometric constraint. Several observations of the planar target are recorded and the constraint is solved using an algebraic approach which gives a closed form solution. \cite{Kwak6094490} calibrate a 2D-LIDAR Camera system using a V-Shaped target and determine the extrinsic parameters by minimizing the distance between the 3D features projected on the image plane and the corresponding image edge features. \cite{Zhou201414} presents a solution to 2D-LIDAR Camera calibration by introducing an additional line to plane constraint to the one used in \cite{Zhang1389752}. The additional constraint is formed by the normal vector detected in the camera frame by checkerboard detection and the scan line direction vector on the checkerboard plane in the LIDAR frame. This additional constraint helps solve for the rotation parameters in a decoupled way. \cite{Ruben7139700} use a target with orthogonal trihedrons using geometric point to back-projected plane constraint described \cite{Naroditsky2011} and the line to plane constraint described in \cite{Zhou201414}. 

The work by \cite{Zhang1389752} was extended to a 3D-LIDAR and Camera system by \cite{unnikrishnan2005} and \cite{Huang2009} using the same geometric constraints. They differ only by the manner they have solved the constrained system of non-linear equations.  \cite{PANDEY2010336} use similar geometric constraints established in \cite{Zhang1389752} to calibrate a 3D-LIDAR and Omni-Directional Camera System. \cite{Zhou201206} present a 3D-LIDAR Camera calibration technique, much similar to the one presented in \cite{Zhou201414} in which the rotation matrix is first determined by line to plane constraint and then a point to plane constraint (similar to ones in \cite{Zhang1389752}, \cite{unnikrishnan2005}, \cite{Huang2009}, \cite{PANDEY2010336}) is used to determine the transformation parameters. \cite{Zhou201810} presents a new method for 3D LIDAR Camera Calibration by adding line to line constraints between lines detected in the 3D-LIDAR and the Camera in addition to the point to plane constraints already exploited in several works described above. 

Most target based 2D/3D-LIDAR Camera extrinsic calibration methods, which use one or more planar surfaces, use checkerboards or ArUco or any other fiducial visual marker for easy detection of planar target in the camera frame. In cases where such markers are not used, V shaped \cite{Kwak6094490}, orthogonal trihedron shaped\cite{Ruben7139700}, perforated targets \cite{butvelodyne} or spherical targets \cite{spherical2018} are employed. 

In this work we use a simple planar surface without any visual markers for external calibration of the 3D LIDAR Camera system. To the best of our knowledge this is the first approach that calibrates a 3D LIDAR Camera system using a simple planar target without any visual marker(s) on them. The geometrical constraint due to back-projected plane has been exploited (in \cite{Naroditsky2011}) to solve the 2D LIDAR-Camera Extrinsic Calibration problem but we haven't come across any work which uses this constraint to calibrate 3D-LIDAR and Camera. We use the geometric constraint due to back-projected plane in our work. Our method requires the planar surface to be bounded by lines and not curves. Although any polygonal plane can be used, we use a square planar surface for the sake of simplicity.

Section \ref{sec: notations} describes the notations used in the paper. Section \ref{sec: featureExtraction} elucidates the feature extraction steps for both LIDAR 
\& Camera data. Section \ref{sec:constraints} lays down the geometrical constraints formed by associated features between LIDAR \& Camera and Section \ref{sec: Optimization} depicts how the aforementioned constraints can form an Optimization problem. A brief system description is given in Section \ref{sec:sysdesc} which follows a Section \ref{sec:experiment} on Experiments and Results. We finally conclude in Section \ref{sec:conclusions}.
\begin{figure}
    \centering
    \includegraphics[scale=0.40]{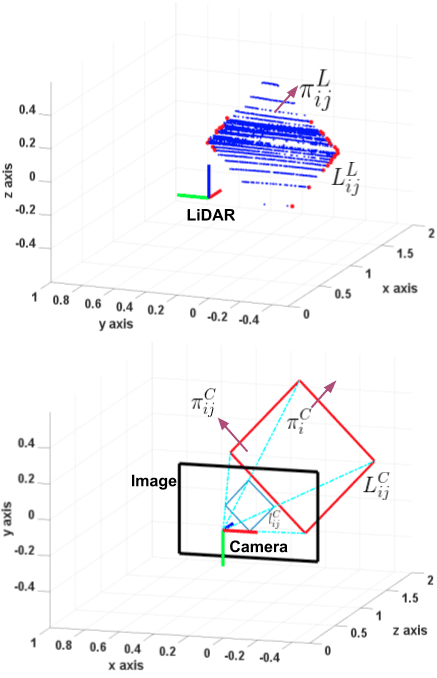}
    \caption{\textbf{Notations:} In LIDAR, the $i^{th}$ pose of the planar target yields planar points $\{P^{L}_{im}\}$ and boundary points $\{Q^L_{ijn}\}$, where $j=\{1, 2, 3, 4\}$, which can be used to estimate $\pi^{L}_{ij}$ and $L^{L}_{ij}$ respectively. In Camera, the $i^{th}$ pose of the planar target yields lines ${l^{C}_{ij}}$, where $j=\{1, 2, 3, 4\}$, which can be used to calculate $\pi^{C}_{i}$, $\pi^{C}_{ij}$ and $L^{C}_{ij}$.}
    \label{fig:schematic}
\end{figure}

\section{Problem Formulation}
For the usual pinhole camera model, the relationship between a homogeneous 3D point, ${P}^L_i$, and its homogeneous image projection $p^C_i$, is given by
\begin{equation}
    p^C_i = K[^CR_L, ^Ct_L]P^L_i.
\end{equation}

The extrinsic parameters that transform the laser coordinate system to that of the camera are captured by $(^CR_L, ^Ct_L)$, where $^CR_L$ is the orthonormal rotation matrix parametrized by the Euler angles $[\phi, \theta, \psi]^\top$ and $t^C_L:=[x,y,z]^\top$ is the translation vector.  The camera intrinsics are captured by the matrix $K$ and is assumed to be known or estimated using Zhang's method \cite{zhang-camera-calib}. We use a planar target that is visible in both camera and lidar frame to establish the geometric constraints that allows us to estimate the rigid body transformation $(^CR_L, ^Ct_L)$ between the two sensors (Figure \ref{fig:schematic}).  

\subsection{Notations}
\label{sec: notations}
The $i^{th}$ pose of the planar target in camera frame is parameterized as $\pi^{C}_{i} = [n^{C}_{i}; d^{C}_{i}]$, where $n^{C}_{i}$ and $d^{C}_{i}$ are the normal to the target's plane and it's distance from the origin of the camera frame of reference. The 4 edges (lines) of the planar target in the image are denoted as $l^{C}_{ij}$. Given the intrinsic camera calibration matrix $K$, the back-projected plane $\pi^{C}_{ij}$ associated with each line $l^{C}_{ij}$ is given by $\pi^{C}_{ij} = [K^{T}l^{C}_{ij}; 0]$ and hence, $n^{C}_{ij} = K^{T}l^{C}_{ij}$ \cite{Hartley2003MVG861369}. The corresponding 3D equation ($L^{C}_{ij}$) of the line $l^{C}_{ij}$  can be determined by computing the intersection of planes $\pi^{C}_{i}$ and $\pi^{C}_{ij}$. The 3D corners $P^{C}_{ij}$ of $\pi^{C}_{i}$ can be obtained by intersection of $\pi^{C}_{i}$ and the back-projected planes $\pi^{C}_{ij}$ of two adjascent lines on image plane. 

For the $i^{th}$ pose of the planar target detected in LIDAR, the normal $n^{L}_{i}$ to $\pi^{L}_{i}$  and centroid $\Bar{P}^{L}_{i}$ can be calculated by using the points $\{P^{L}_{im}\}$ on $\pi^{L}_{i}$. Similarly, using points $\{Q^{L}_{ijn}\}$ on $L^{L}_{ij}$, the direction $d^{L}_{ij}$ of $L^{L}_{ij}$ and the centroid $\Bar{Q}^{L}_{ij}$ can be obtained. 

\subsection{Feature Extraction}
\label{sec: featureExtraction}
The line ($l^{C}_{ij}$) in the image is detected using Line Segment Detector \cite{LSD4731268} available in OpenCV \cite{opencvlibrary}. For a four sided polygon shown in Figure \ref{fig:schematic} the intersection of adjacent lines $l^{C}_{ij}$ provide the corner points ($p^{C}_{ij}$) in the image plane. The known dimensions of the planar target is used to solve a PnP problem (using OpenCV) and $\pi^{C}_{i}$ is obtained. To summarize, the features obtained from the image are the edges $l^{C}_{ij}$ of the planar target, the corners $p^{C}_{ij}$ and the plane $\pi^{C}_{i}$. 

\begin{figure}[!ht]
\centering
    \includegraphics[scale=0.30]{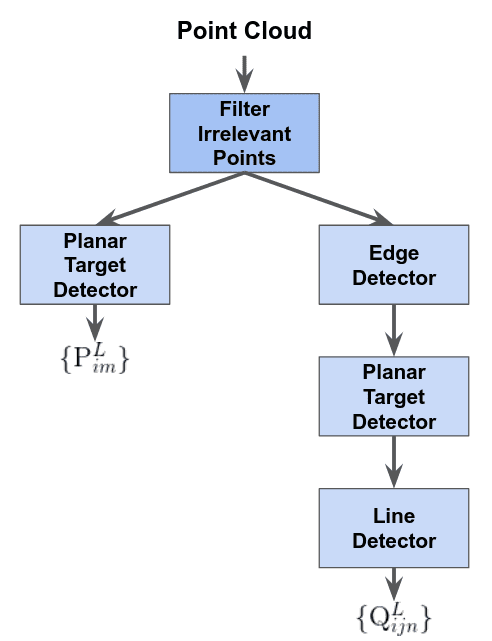}
    \caption{Pipeline for extraction of Planar Points $\{P^{L}_{im}\}$ and the points on the $j^{th}$ edge $\{Q^{L}_{ijn}\}$ in LIDAR}
    \label{fig:pointcloudprocessing}
\end{figure}

\begin{figure}[!tbp]
  \centering
  \subfloat[Features in Image.]{\includegraphics[width=0.23\textwidth]{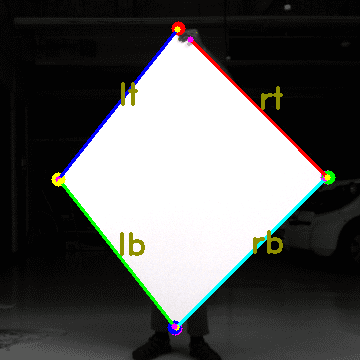}\label{fig: image_lines}}
  \hfill
  \subfloat[Features in PointCloud.]{\includegraphics[width=0.23\textwidth]{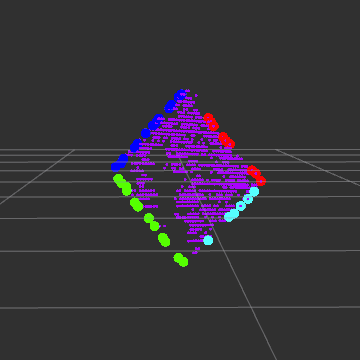}\label{fig: LIDAR_lines}}
  \caption{Result of Feature Detection in Image and LIDAR.}
  \label{fig: features}
\end{figure}

The block diagram in Figure \ref{fig:pointcloudprocessing} shows the feature detection process for point-cloud data from LIDAR. The point cloud data is filtered to remove points which have very low probability of lying in the common field of view of the camera and the LIDAR.  For planar target detection we apply plane segmentation on the filtered cloud to get the points $\{P^{L}_{im}\}$ lying on the planar target using a RANSAC \cite{RANSAC} based method available in the Point Cloud Library (PCL) \cite{PCL}. 

We follow a three step approach to detect the points lying along the boundaries of the planar target. First we apply an edge detection algorithm \cite{Levinson201306} to the filtered point-cloud to obtain an edge point-cloud. Second, we apply plane segmentation to the edge point-cloud and obtain a point-cloud formed by points which lie on the planar target's boundaries. Next, we apply line detection to the point-cloud formed by the boundary points of the planar target. The RANSAC base line detection algorithm (available in \cite{PCL}) fits 4 lines to the point-cloud formed by the target's boundaries and classifies each point $\{Q^{L}_{ijn}\}$ as associated to one of the four detected lines. As scan lines are parallel to the LIDAR's horizontal plane, any edge parallel to the LIDAR's horizontal plane will be difficult to detect. Therefore edge/line detection in LIDAR is easier when the planar target is held in a way similar to Figure \ref{fig: features}. In summary, the features obtained from the LIDAR data are the points on the plane $\{P^{L}_{im}\}$ and the points on the respective edges $\{Q^{L}_{ijn}\}$.


\subsection{Constraints}
\label{sec:constraints}
Using the features extracted in the previous section, several geometric constraints can be established which involve the rotation $^CR_{L}$ and the translation $^Ct_{L}$ between the LIDAR and Camera centers. Two such constraints pertinent to this work are discussed below.
\subsubsection{Point to Plane Constraint}
\label{sec: p2pC}
Given the $i^{th}$ pose of the planar target, a point $P^{L}_{im}$ lying on the planar target in the LIDAR coordinate frame and the plane parameters $\pi^{C}_{i}$ satisfies the geometric constraint given in Equation \ref{eqn:point2plane}.
\begin{equation}
    n^{C}_{i}.(^{C}R_{L}P^{L}_{im} + ^{C}t_{L} - d^{C}_{i}) = 0
    \label{eqn:point2plane}
\end{equation}
As declared earlier, $\pi^{C}_{i} = [n^{C}_{i}; d^{C}_{i}]$. This is the standard point-to-plane projection constrain used by other target based methods as well \cite{Zhang1389752}.
\subsubsection{Point to Back-projected Plane Constraint}
\label{sec: p2BpC}
This is the additional constrain that we introduce in this paper that helps in refining the calibration parameters. Given the $i^{th}$ pose of the planar target, a point $Q^{L}_{ijn}$ lying on the $j^{th}$ edge of the planar target in the LIDAR coordinate frame and a corresponding line $l^{C}_{ij}$ in the image satisfy the following geometric constraint:
\begin{equation}
    n^{C}_{ij}.(^{C}R_{L}Q^{L}_{ijn} + ^{C}t_{L}) = 0
    \label{eqn:point2backprojplane}
\end{equation}
Where $n^{C}_{ij} = K^{T}l^{C}_{ij}$ is the normal to the back-projected plane formed by the camera center and the line $l_{ij}$ (Fig \ref{fig:schematic}). 

Equations \ref{eqn:point2plane} and \ref{eqn:point2backprojplane} are equations of the planar target and the back-projected plane respectively, in the \textit{Hessian} normal form, in the camera frame.
 
\subsection{Optimization}
\label{sec: Optimization}
Cost functions for solving an optimization problem are formulated in the following ways using the geometrical constraints given in Equation \ref{eqn:point2plane} and \ref{eqn:point2backprojplane}:
\subsubsection{Cost Function from Point to Plane Constraint}
The cost function formed by Point to Plane Constraint is given in Equation \ref{eqn: CostFn1}.
\begin{multline}
\label{eqn: CostFn1}
P_1 = \sum_{i=1}^{M}\frac{1}{p_i}\sum_{m=1}^{p_{i}} \norm{(n^{C}_{i})^{T}(^{C}R_{L}P^{L}_{im} + ^{C}t_{L} - d^{C}_{i})}^{2}
\end{multline}
Here, $p_i$ is the number of LIDAR points lying on the planar target in the $i^{th}$ observation and $M$ is the total number of observations.
To obtain an estimate $[^C\Tilde{R}_L, ^C\Tilde{t}_L]$, Equation \ref{eqn: CostFn1} needs to be minimized with respect to $[^CR_L, ^Ct_L]$.
\begin{equation}
\label{eqn: CostFn1Minimization}
    [^C\Tilde{R}_L, ^C\Tilde{t}_L] = \argmin_{[^CR_L, ^Ct_L]} P_1
\end{equation}

\subsubsection{Cost Function from Point to Back-projected Plane Constraint}
The cost function formed by Point to Back-projected Plane Constraint is given in Equation \ref{eqn: CostFn2}.
\begin{multline}
\label{eqn: CostFn2}
P_2 = \sum_{i=1}^{N} \sum_{j=1}^{4} \frac{1}{q_{ij}} \sum_{m=1}^{q_{ij}} \norm{(n^{C}_{ij})^{T}(^{C}R_{L}Q^{L}_{ijm} + ^{C}t_{L})}^{2}
\end{multline}
Here $q_{ij}$ is the number of points lying on the $j^{th}$ line in the $i^{th}$ observation and $N$ is the number of observations. To obtain an estimate $[^C\Tilde{R}_L, ^C\Tilde{t}_L]$, Equation \ref{eqn: CostFn2} needs to be minimized with respect to $[^CR_L, ^Ct_L]$.
\begin{equation}
\label{eqn: CostFn2Minimization}
    [^C\Tilde{R}_L, ^C\Tilde{t}_L] = \argmin_{[^CR_L, ^Ct_L]} P_2
\end{equation}

We have found experimentally that the minimization problem given in Equation \ref{eqn: CostFn2Minimization} is affected by the quality of initialization, especially of the rotational variables. The minimization problem given in Equation \ref{eqn: CostFn1Minimization} converges to correct rotation parameters but it consistently converges to in-correct translation parameters, irrespective of the value of initialization. Therefore, we solve Equation \ref{eqn: CostFn1Minimization} first and then use the results obtained to initialize the problem given in Equation \ref{eqn: CostFn2Minimization} before solving it. We use the Ceres \cite{ceres-solver} non-linear least squares solver in our work.

As mentioned in \cite{PANDEY2010336}, we need at-least 3 non-coplanar views to solve the optimization problem formed by Equation \ref{eqn: CostFn1Minimization}. This is ensured by monitoring the value of $\pi^{C}_{i}$ which is obtained in the feature detection step. 

The \textit{Point to Back-projected Plane} constraint given in Equation \ref{eqn:point2backprojplane} is equivalent to the line correspondence equation given in \cite{Hartley2003MVG861369} (2004, p. 180). The solution to such a system of equation is given by the DLT-Lines method and requires at least 6 noise free line correspondences \cite{PribylZC16a} between the LIDAR and camera views. Since our planar target has 4 sides , ideally, we need at least 2 distinct views to solve this system. 

In practice it is advisable to get sufficient number of views, for both the constraints, because that ensures diversity of measurement and robustness of the solution. 

The overall algorithm is summarized below in Algorithm \ref{Algorithm}:

\begin{algorithm}
\KwIn{$N$ pairs of LIDAR and Camera Data containing the Planar Target in Field of View}
\KwOut{Rigid transformation parameters $[R^{C}_{L}, t^{C}_{L}]$ from LIDAR to Camera}
\textbf{Algorithm Steps}\\
\textbf{1.} Collect $N$ pairs of LIDAR and Camera Data with Planar Target in Field of View.\\
\textbf{2.} Detect the edges $\{l^{C}_{ij}\}$ of the Planar Target in Camera Images and use them to detect the Planar Target $\pi^{C}_{i}$ in Camera Frame.\\
\textbf{3.} Detect points $\{P^{L}_{im}\}$ lying on the Planar Target in LIDAR data.\\
\textbf{4.} Detect points $\{Q^{L}_{ijn}\}$ lying on the edges of the Planar Target in LIDAR data.\\
\textbf{5.} Use the detected features to form the Cost Functions \ref{eqn: CostFn1} and \ref{eqn: CostFn2}.\\
\textbf{6.} Solve the minimization problem given in Equation \ref{eqn: CostFn1Minimization}.\\
\textbf{7.} Use the solution obtained above to initialize and solve the minimization problem given in Equation \ref{eqn: CostFn2Minimization}.
\caption{\bf Extrinsic Calibration using Plane and Line Correspondences}
\label{Algorithm}
\end{algorithm}

\section{System Description}
\label{sec:sysdesc} 
Our setup consists of an Ouster OS1 64 Channel LIDAR, a Velodyne VLP-32 LIDAR and two Basler cameras which come as Karmin2 Stereo rig such that the factory stereo calibration is known and can be used as one of the methods to test the veracity of the result produced by our algorithm. 

The data from all the sensors come at 10 Hz. The sensors are not hardware synchronized but an approximate software time synchronization \footnote{using ROS \cite{ROS} message filters} is followed while collecting data from the sensors.
\begin{figure}[!ht]
\centering
    \includegraphics[scale=0.10]{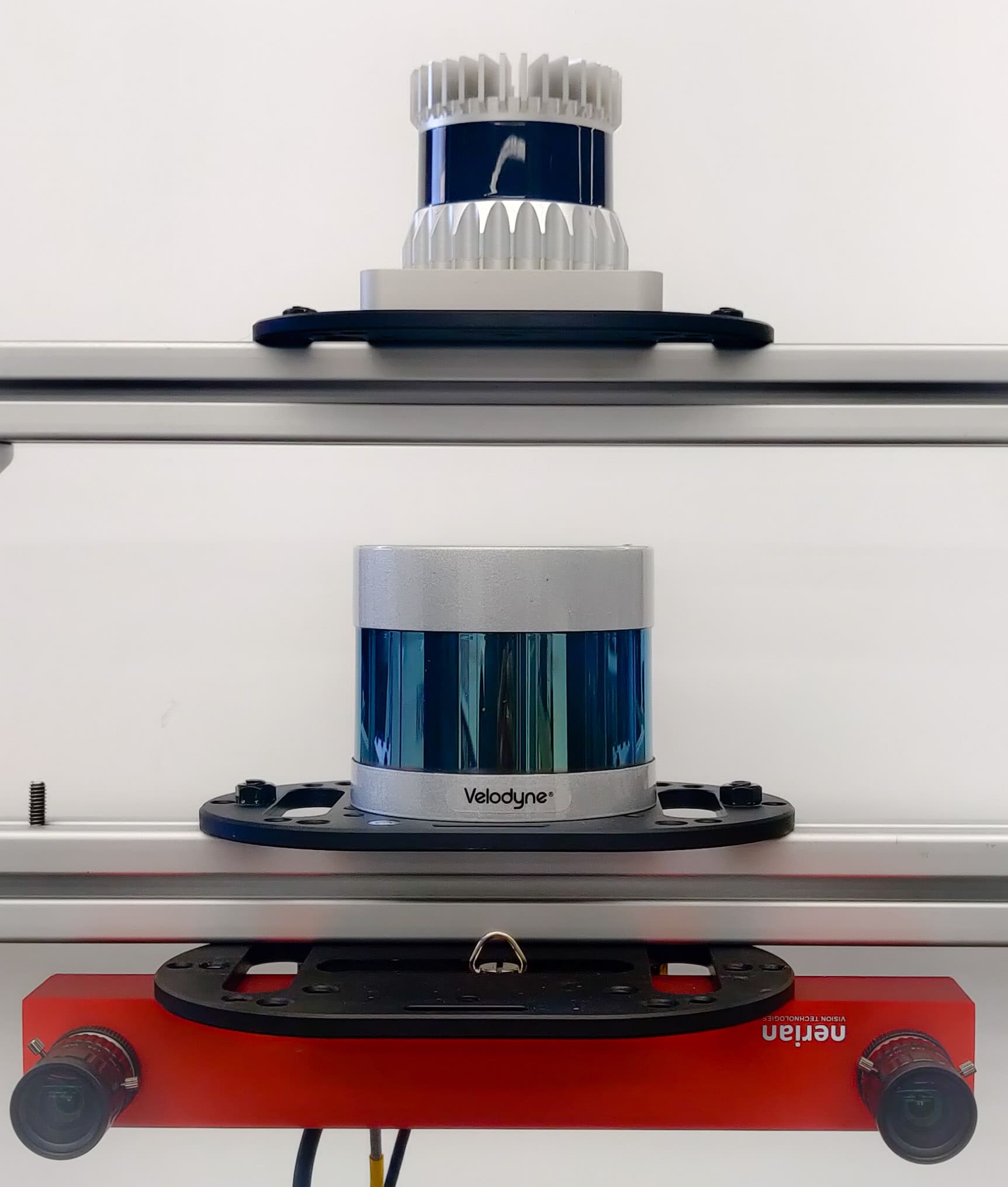}
    \caption{\textbf{Sensors (From Top to Bottom):} Ouster OS1 64 Channel LIDAR, Velodyne VLP-32 LIDAR and Karmin2 Stereo (Basler) Camera}
    \label{fig: sensorsuite}
\end{figure}
\section{Experiments and Results}
\label{sec:experiment}
In this section we use our algorithm on real world data collected using sensors shown in Figure \ref{fig: sensorsuite}. The four possible LIDAR-Camera setups are Velodyne VLP-32 LIDAR with left Basler Camera, Velodyne VLP-32 LIDAR with right Basler Camera, Ouster LIDAR with left Basler Camera and Ouster LIDAR with right Basler Camera. 

\subsection{Data Collection}
\label{sec:datacollection} 
As our sensors are not time synchronized, it is advisable to collect data frames only when the planar target is at rest. We have a motion detection module which triggers feature detection only when the target is at rest for a user-defined duration.
\subsubsection{Data Required for Solving Point to Plane Constraint}
\label{sec: datacollec1}
The data required for solving the minimization problem (Equation \ref{eqn: CostFn1Minimization}) formed by the constraint given in Equation \ref{eqn:point2plane} are: the points lying on the planar target $\{P^{L}_{im}\}$,  the normal vector from the camera to the planar surface in camera frame $n^{C}_{i}$ and the vector from the origin of the camera frame to the origin of the planar target $d^{C}_{i}$ in the camera frame. Data is collected by placing the Planar Target at several positions and orientations infront of the sensor suite. 

The Ouster OS1 LIDAR was found to be noisier than the Velodyne VLP-32 LIDAR, because the plane segmentation algorithm returned fewer planar points when the same RANSAC parameters were used for both the LIDARs. On an average, we received 2600 planar points from Velodyne VLP-32 while we got 1200 planar points from Ouster OS1, despite the latter being a 64 channel LIDAR.

\subsubsection{Data Required for Solving Point to Back-Projected Plane Constraint}
The data required for solving the minimization problem (Equation \ref{eqn: CostFn2Minimization}) formed by the constraint given in Equation  \ref{eqn:point2backprojplane} are: the points lying on the edges of the planar target in the LIDAR frame $\{Q^{L}_{ijm}\}$ and the normal $n^{C}_{ij}$ to the back-projected plane formed by line $l^{C}_{ij}$ detected in the camera frame. 

While collecting data, it is essential to note that the orientation of the target is well within the thresholds which allows good edge detection. Since Velodyne VLP-32 has a non-linear distribution of scan rings about y-axis, the rings are concentrated near the origin and sparser as one pitches up or down (can be seen in Figure \ref{fig: vlp_nerian}). Therefore, the planar pattern has to be held at a suitable height so that sufficient points from its surface and edges can be collected.

We collect about 30 frames from each sensor pair with planar target at different view points and not all of the frames were used in both the minimization problems. 

\subsection{Results}
\label{sec:Results} 
\begin{figure}[!ht]
  \centering
  \subfloat[Results with method presented in \cite{PANDEY2010336} for Left Camera and Ouster LIDAR]{\includegraphics[width=0.23\textwidth]{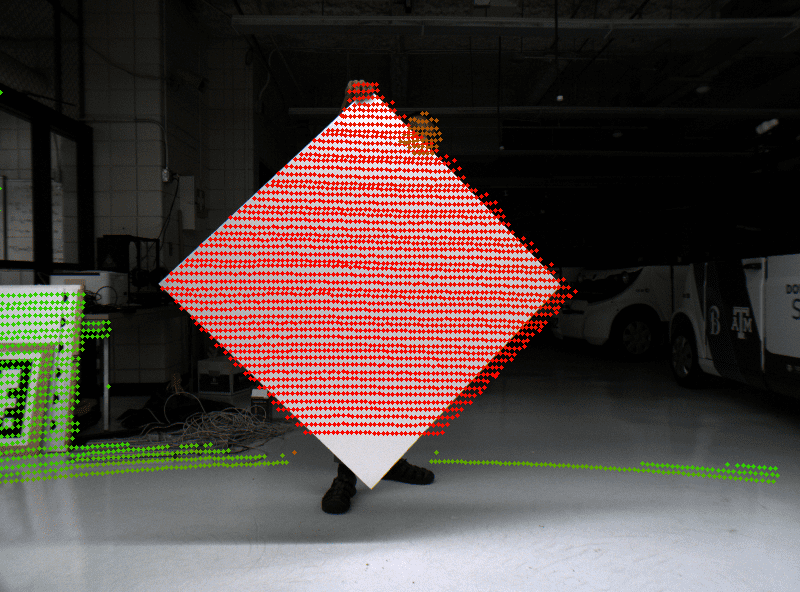}\label{fig: left_os_gp}}
  \quad
  \subfloat[Results with our method for Left Camera and Ouster LIDAR]{\includegraphics[width=0.23\textwidth]{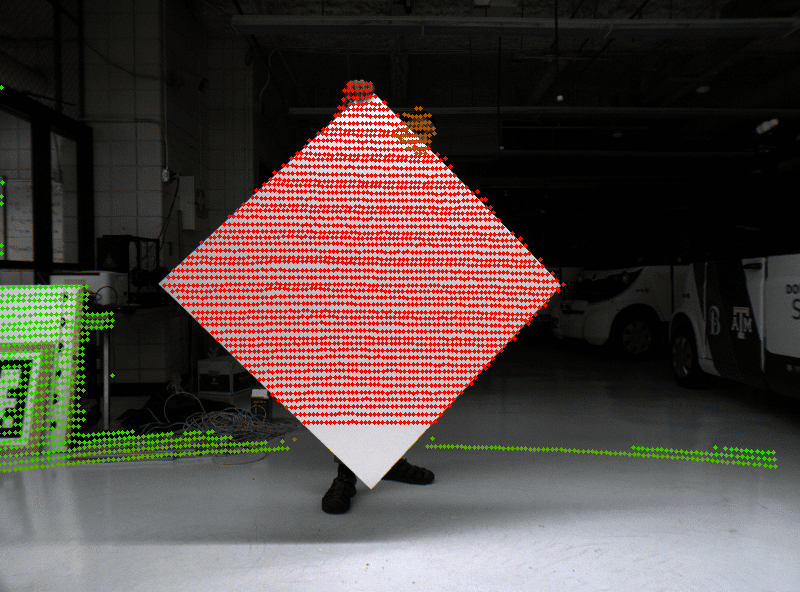}\label{fig: left_os_mn}}\\
  \subfloat[Results with method presented in \cite{PANDEY2010336} for Right Camera and Ouster LIDAR]{\includegraphics[width=0.23\textwidth]{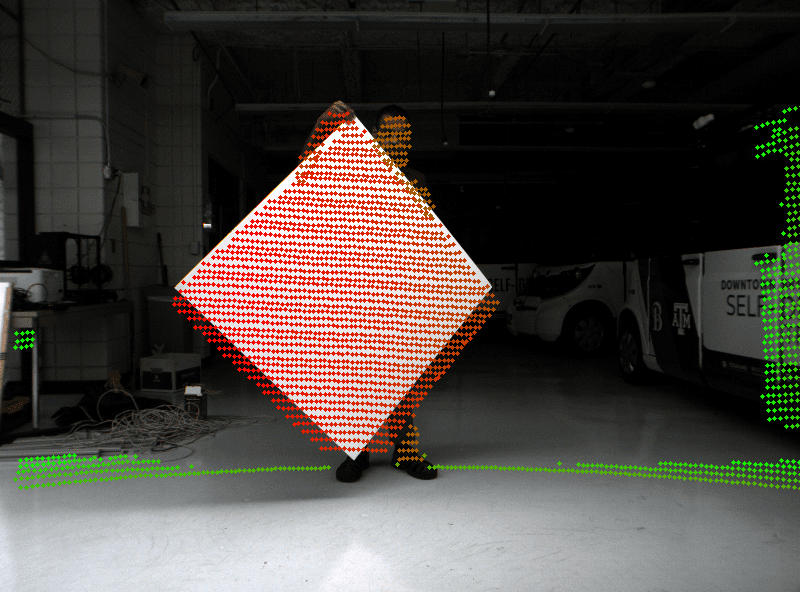}\label{fig: right_os_gp}}
  \quad
  \subfloat[Results with our method for Right Camera and Ouster LIDAR]{\includegraphics[width=0.23\textwidth]{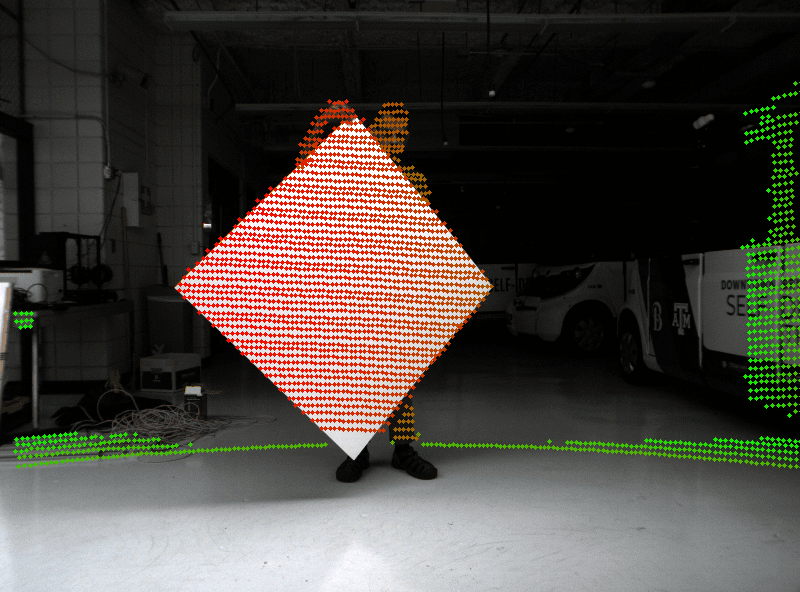}\label{fig: right_os_mn}}
  \caption{Comparing \cite{PANDEY2010336} with our method for a setup comprising an Ouster 64 Channel LIDAR \& Left (Figures \ref{fig: left_os_gp} and \ref{fig: left_os_mn}) and Right (Figures \ref{fig: right_os_gp} and \ref{fig: right_os_mn}) Basler Cameras of the sensor suite shown in Figure \ref{fig: sensorsuite}}
  \label{fig: ouster_nerian}
\end{figure}

\begin{figure}[!ht]
  \centering
  \subfloat[Results with method presented in \cite{PANDEY2010336} for Left Camera and Velodyne VLP-32 LIDAR]{\includegraphics[width=0.23\textwidth]{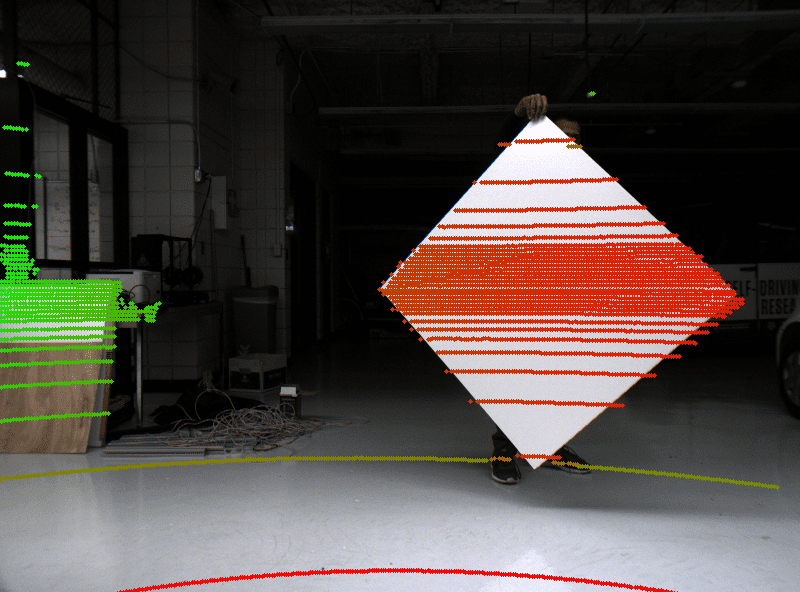}\label{fig: left_VLP_gp}}
  \quad
  \subfloat[Results with our method for Left Camera and Velodyne VLP-32 LIDAR]{\includegraphics[width=0.23\textwidth]{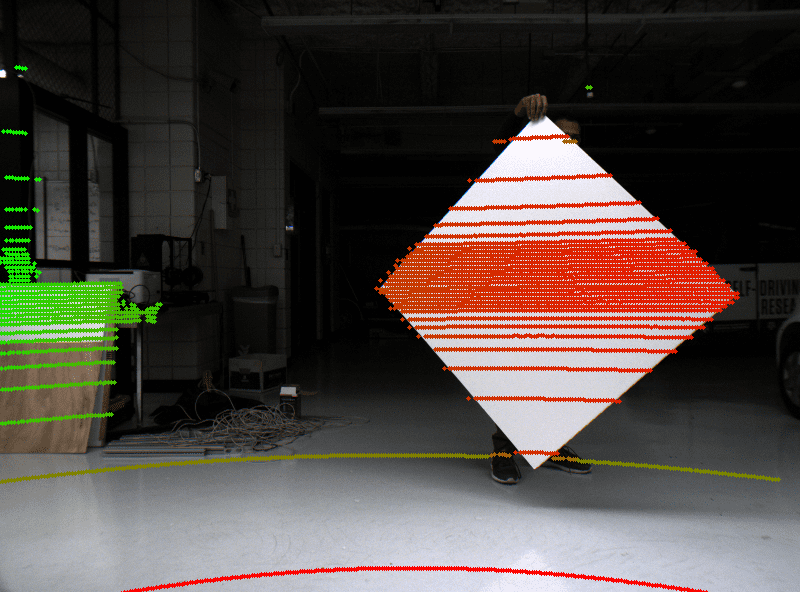}\label{fig: left_vlp_mn}}\\
    \subfloat[Results with method presented in \cite{PANDEY2010336} for Right Camera and Velodyne VLP-32 LIDAR]{\includegraphics[width=0.23\textwidth]{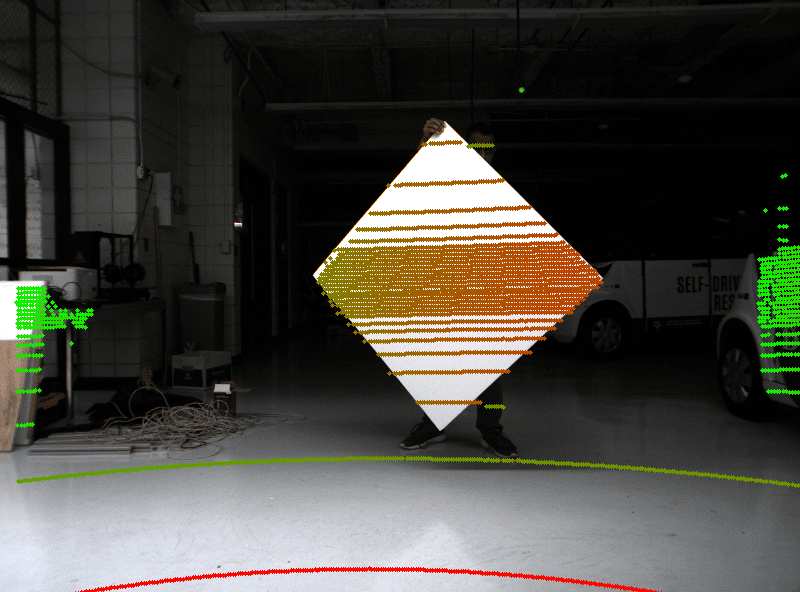}\label{fig: right_VLP_gp}}
  \quad
  \subfloat[Results with our method for Right Camera and Velodyne VLP-32 LIDAR]{\includegraphics[width=0.23\textwidth]{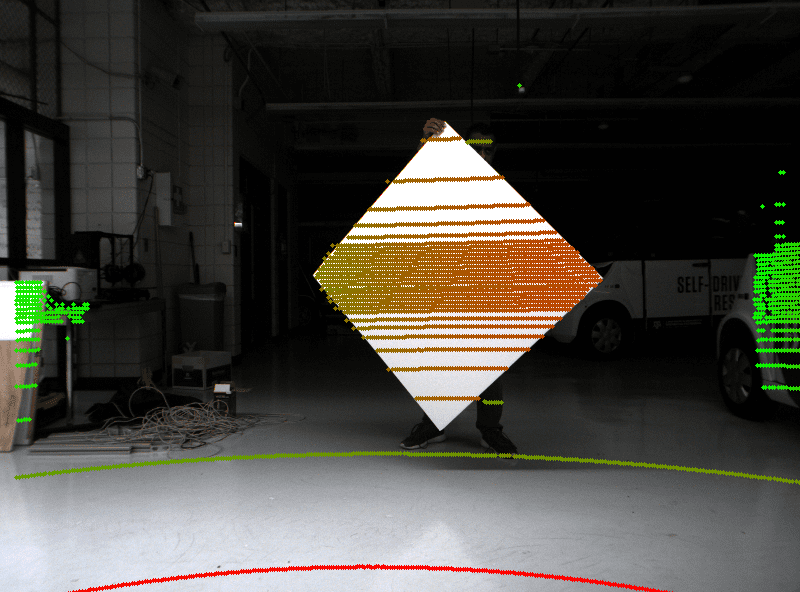}\label{fig: right_vlp_mn}}
  \caption{Comparing \cite{PANDEY2010336} with our method for a setup comprising a Velodyne VLP-32 LIDAR \& Left (Figures \ref{fig: left_VLP_gp} and \ref{fig: left_vlp_mn}) and Right (Figures \ref{fig: right_VLP_gp} and \ref{fig: right_vlp_mn}) Basler Cameras of the sensor suite shown in Figure \ref{fig: sensorsuite}}
  \label{fig: vlp_nerian}
\end{figure}

\begin{figure}[!ht]
  \centering
  \subfloat[Average line re-projection error of \textcolor{red}{7.01005} with method presented in \cite{PANDEY2010336}.]{\includegraphics[width=0.23\textwidth]{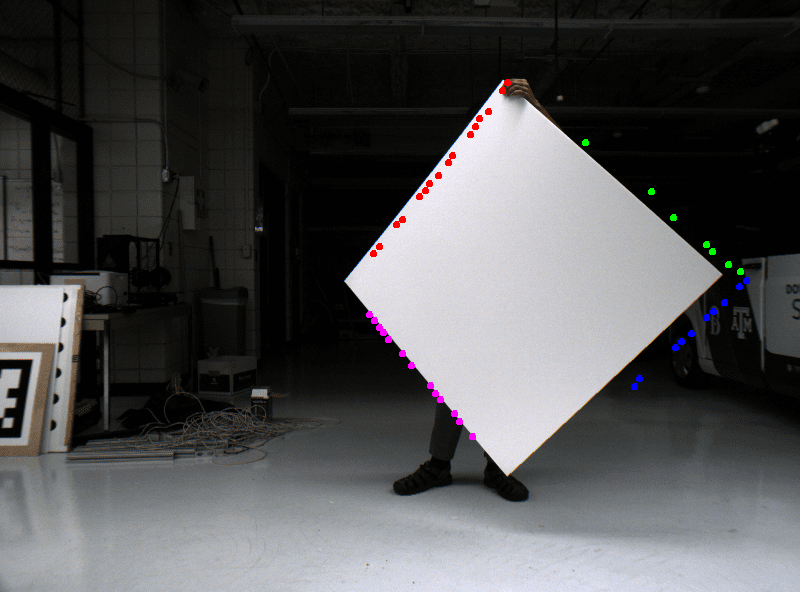}\label{fig: left_os_gp_line}}
  \quad
  \subfloat[Average line re-projection error of \textcolor{blue}{1.88829} with our method.]{\includegraphics[width=0.23\textwidth]{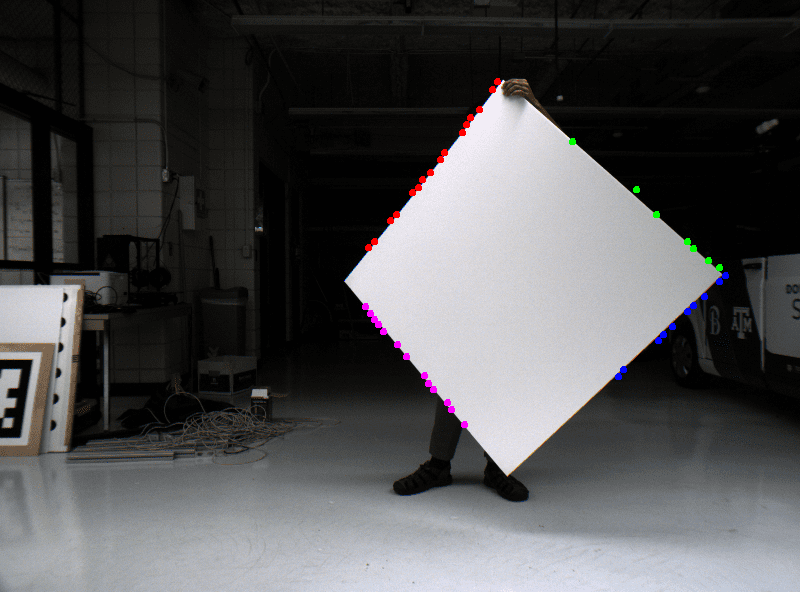}\label{fig: left_os_mn_line}}\\
  \subfloat[Average line re-projection error of \textcolor{red}{9.64533} with method presented in \cite{PANDEY2010336}.]{\includegraphics[width=0.23\textwidth]{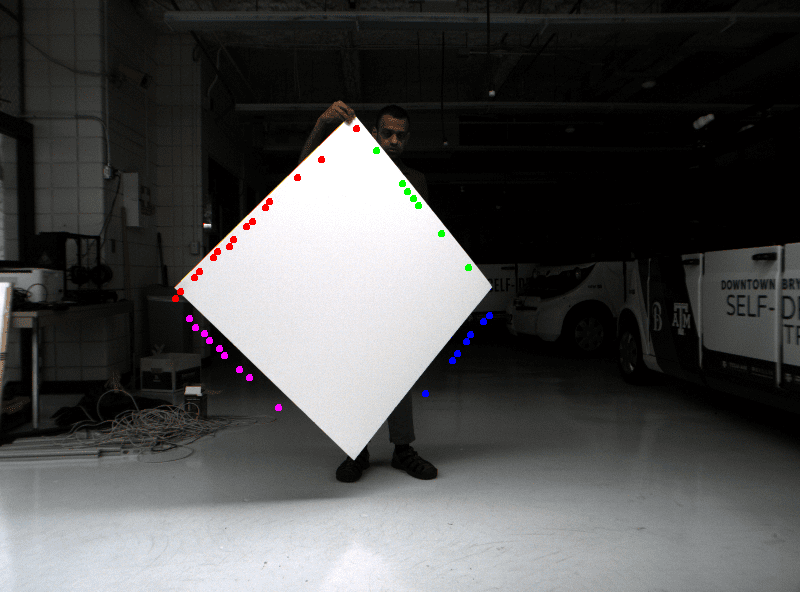}\label{fig: right_os_gp_line}}
  \quad
  \subfloat[Average line re-projection error of \textcolor{blue}{1.84383} with our method.]{\includegraphics[width=0.23\textwidth]{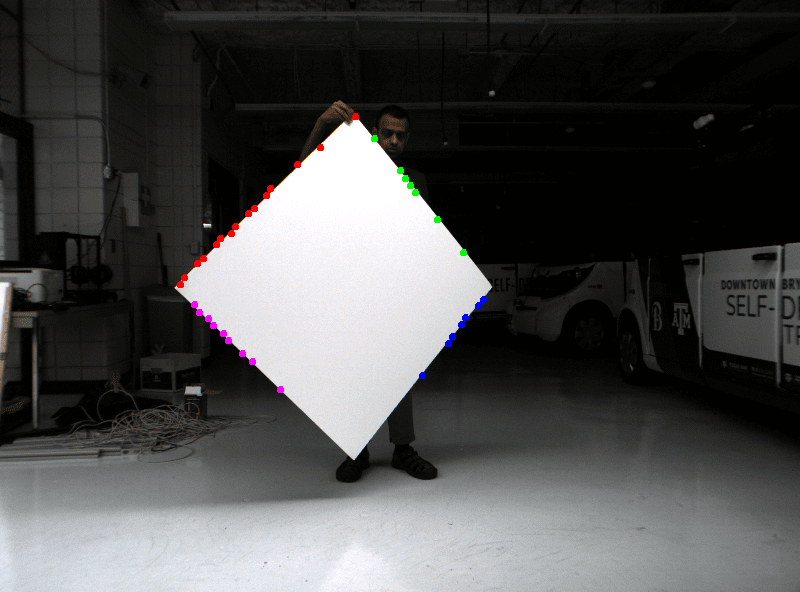}\label{fig: right_os_mn_line}}
  \caption{Comparing \cite{PANDEY2010336} with our method for a setup comprising an Ouster 64 Channel LIDAR \& Left (Figures \ref{fig: left_os_gp_line} and \ref{fig: left_os_mn_line}) and Right (Figures \ref{fig: right_os_gp_line} and \ref{fig: right_os_mn_line}) Basler Cameras of the sensor suite shown in Figure \ref{fig: sensorsuite}}
  \label{fig: ouster_nerian_lines}
\end{figure}

\begin{figure}[!ht]
  \centering
  \subfloat[Average line re-projection error of \textcolor{red}{3.43154} with method presented in \cite{PANDEY2010336}.]{\includegraphics[width=0.23\textwidth]{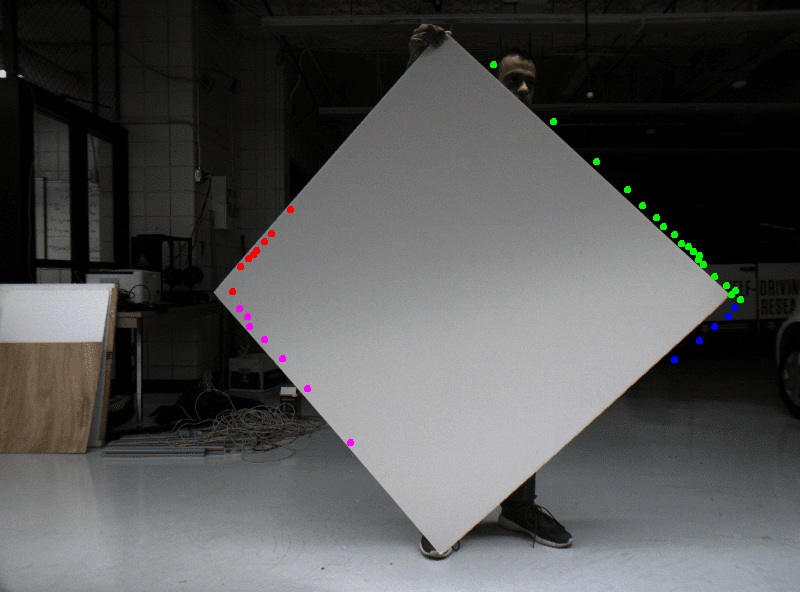}\label{fig: left_VLP_gp_line}}
  \quad
  \subfloat[Average line re-projection error of \textcolor{blue}{1.95631} with our method.]{\includegraphics[width=0.23\textwidth]{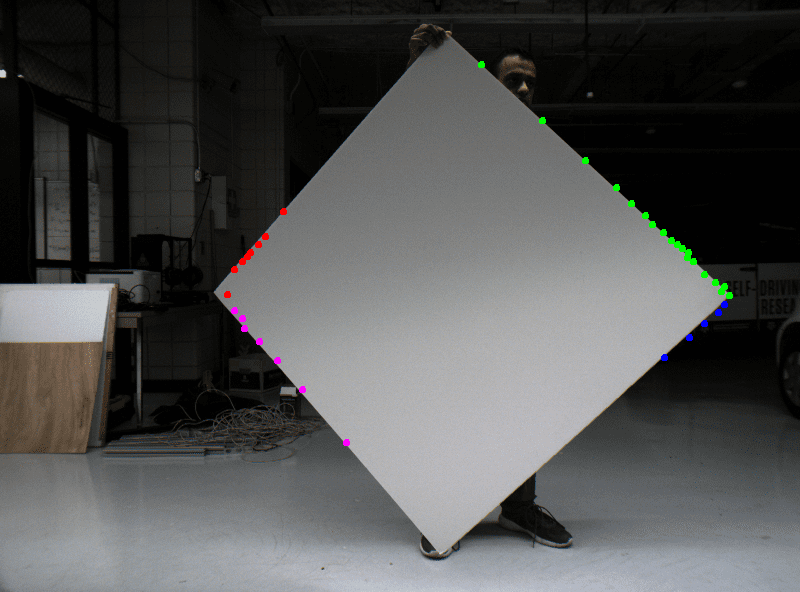}\label{fig: left_vlp_mn_line}}\\
    \subfloat[Average line re-projection error of \textcolor{red}{3.0122} with method presented in \cite{PANDEY2010336}.]{\includegraphics[width=0.23\textwidth]{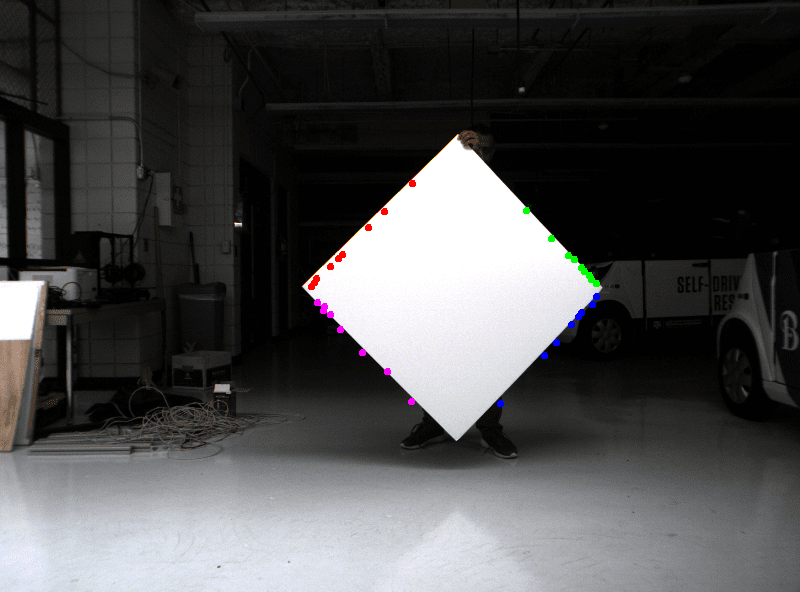}\label{fig: right_VLP_gp_line}}
  \quad
  \subfloat[Average line re-projection error of \textcolor{blue}{2.07706} with our method.]{\includegraphics[width=0.23\textwidth]{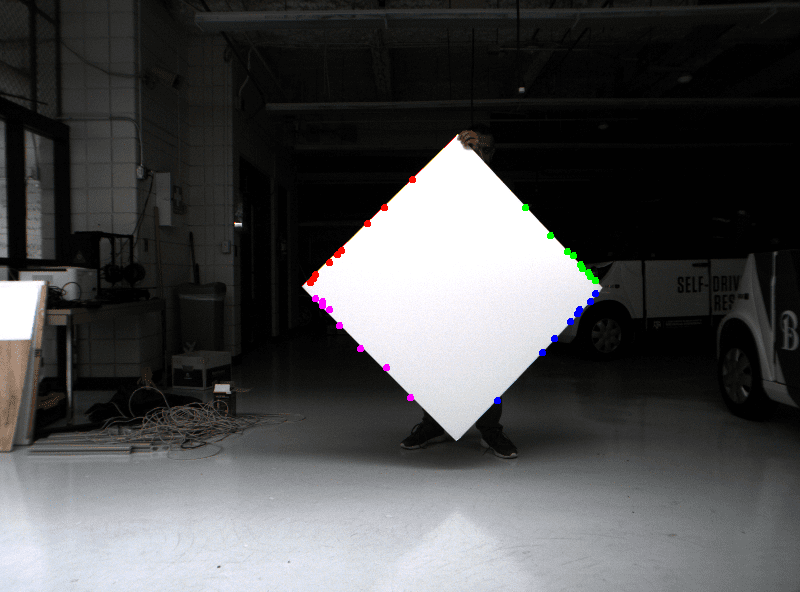}\label{fig: right_vlp_mn_line}}
  \caption{Comparing \cite{PANDEY2010336} with our method for a setup comprising a Velodyne VLP-32 LIDAR \& Left (Figures \ref{fig: left_VLP_gp_line} and \ref{fig: left_vlp_mn_line}) and Right (Figures \ref{fig: right_VLP_gp_line} and \ref{fig: right_vlp_mn_line}) Basler Cameras of the sensor suite shown in Figure \ref{fig: sensorsuite}}
  \label{fig: vlp_nerian_lines}
\end{figure}

We compare the performance of our method with \cite{PANDEY2010336} for all the experimental setups. For calibrating using \cite{PANDEY2010336} we use a $6\times9$ checkerboard with each checker sized 0.061 x 0.061 $m^2$. In the absence of ground truth we verify our result by using the estimated parameters \begin{enumerate*} [label=\itshape\alph*\upshape)]
\item to project LIDAR points on camera image as shown in Figures \ref{fig: ouster_nerian} \& \ref{fig: vlp_nerian} , \item to project points lying on the edges of the planar target in LIDAR frame on the Camera image and calculate the average line re-projection errors\footnote{The average distance between $\{l^{C}_{ij}\}$ and $\{Q^{L}_{ijn}\}$ projected on the image using the estimated $[^CR_{L}, ^Ct_{L}]$} as shown in Figures \ref{fig: ouster_nerian_lines} \& \ref{fig: vlp_nerian_lines} and \item to compare it against the factory stereo calibration as shown in Tables \ref{table: stereo_os} \& \ref{table: stereo_vlp}\footnote{We use the estimated $T^{C_1}_L$ and $T^{C_2}_L$ and compare $T^{C_1}_{L}(T^{C_2}_{L})^{-1}$ with the given factory stereo calibration $T^{C_1}_{C_2}$}
\end {enumerate*}.

Comparing Figures \ref{fig: left_os_gp}, \ref{fig: right_os_gp}, \ref{fig: left_VLP_gp}, \ref{fig: right_VLP_gp} with \ref{fig: left_os_mn}, \ref{fig: right_os_mn}, \ref{fig: left_vlp_mn}, \ref{fig: right_vlp_mn} respectively, the projection of LIDAR points with extrinsics estimated using \cite{PANDEY2010336} overshoot the planar target in the image plane but the projection shows perfect alignment when using our method, for all sensor pairs.

Comparing Figures \ref{fig: left_os_gp_line}, \ref{fig: right_os_gp_line}, \ref{fig: left_VLP_gp_line}, \ref{fig: right_VLP_gp_line} with \ref{fig: left_os_mn_line}, \ref{fig: right_os_mn_line}, \ref{fig: left_vlp_mn_line}, \ref{fig: right_vlp_mn_line} respectively we can see that the average line re-projection errors are higher when using \cite{PANDEY2010336} as compared to our method, for all sensor pairs.

\begin{table}[!ht]
\setlength{\tabcolsep}{2.25pt}
\centering
\begin{tabular}{|c | c c c c c c|} 
 \hline
 Method & $\alpha^{\circ}_{err}$ & $\beta^{\circ}_{err}$ &$ \gamma^{\circ}_{err}$ & $X_{err}$ $[m]$ & $Y_{err}$ $[m]$ &  $Z_{err}$ $[m]$\\ [0.5ex] 
 \hline
 \cite{PANDEY2010336} & \textcolor{red}{-0.45112} & \textcolor{blue}{3.4151} & \textcolor{red}{-0.52642} & \textcolor{red}{-0.06923} & \textcolor{red}{-0.01183} & \textcolor{red}{0.01334}\\  [1ex] 
  \hline 
  Proposed & \textcolor{blue}{-0.063075} & \textcolor{red}{4.5844} & \textcolor{blue}{0.19586} & \textcolor{blue}{0.00134} & \textcolor{blue}{-0.00736} & \textcolor{blue}{0.00626}\\  [1ex] 
  \hline
\end{tabular}
\caption{Errors with respect to factory stereo calibration for Ouster 64 Channel LIDAR and the stereo rig}
\label{table: stereo_os}
\end{table}

\begin{table}[!ht]
\setlength{\tabcolsep}{2.25pt}
\centering
\begin{tabular}{|c | c c c c c c|} 
 \hline
 Method & $\alpha^{\circ}_{err}$ & $\beta^{\circ}_{err}$ &$ \gamma^{\circ}_{err}$ & $X_{err}$ $[m]$ & $Y_{err}$ $[m]$ &  $Z_{err}$ $[m]$\\ [0.5ex] 
 \hline
 \cite{PANDEY2010336} & \textcolor{red}{-1.1072} & \textcolor{blue}{3.4868} & \textcolor{red}{-0.16217} & \textcolor{red}{-0.05363} & \textcolor{red}{-0.02342} & \textcolor{red}{0.00537}\\  [1ex] 
  \hline 
  Proposed & \textcolor{blue}{-0.020025} & \textcolor{red}{4.4823} & \textcolor{blue}{0.41197} & \textcolor{blue}{0.00398} & \textcolor{blue}{0.00268} & \textcolor{blue}{-0.00205}\\  [1ex] 
  \hline
\end{tabular}
\caption{Errors with respect to factory stereo calibration for Velodyne VLP-32 LIDAR and the stereo rig}
\label{table: stereo_vlp}
\end{table}

The errors in the estimated parameters with respect to the factory stereo calibration for Karmin2 Stereo Vision system for Ouster OS1 and Velodyne VLP-32 LIDARs are presented in Tables \ref{table: stereo_os} and \ref{table: stereo_vlp} respectively. For both the LIDARs, our method shows better performance for all the estimated parameters except for the angular error in $\beta$.

If we compare the difference in average line re-projection errors between the method in \cite{PANDEY2010336} and our method for both Ouster OS1 and Velodyne VLP-32 in Figures \ref{fig: ouster_nerian_lines} and \ref{fig: vlp_nerian_lines} respectively, we find that it is much higher in the case of Ouster OS1 than Velodyne VLP-32. It is because the Ouster OS1 LIDAR is noisier than Velodyne VLP-32 according to our experiments described in Section \ref{sec: datacollec1}.

\section{Conclusions and Future Work}
\label{sec:conclusions}
In this work we showed how two different geometric constraints can be used in series to determine the extrinsic calibration of a 3D-LIDAR Camera system and extensively used our algorithm across different sensors. Real world experiments show that the method performed better than one presented in \cite{PANDEY2010336} which is a widely used and easily implementable method to do LIDAR-Camera extrinsic calibration.

In the future we want to improve and make the feature detection step robust, initialize the non-linear least squares with the closed form solutions obtained from linear least squares, use the average line re-projection error as weight for calibration refinement algorithms, integrate an inexpensive IMU to our sensor suite and jointly calibrate the system using Graph Based methods \cite{Owens_msg-cal:multi-sensor}. In addition to these, we also intend to implement algorithms which can detect change in calibration due to environmental or other factors when the system is operating in real time.
\bibliographystyle{IEEEtran}
\bibliography{bibexpendable}

\begin{thebibliography}{10}
\providecommand{\url}[1]{#1}
\csname url@samestyle\endcsname
\providecommand{\newblock}{\relax}
\providecommand{\bibinfo}[2]{#2}
\providecommand{\BIBentrySTDinterwordspacing}{\spaceskip=0pt\relax}
\providecommand{\BIBentryALTinterwordstretchfactor}{4}
\providecommand{\BIBentryALTinterwordspacing}{\spaceskip=\fontdimen2\font plus
\BIBentryALTinterwordstretchfactor\fontdimen3\font minus
  \fontdimen4\font\relax}
\providecommand{\BIBforeignlanguage}[2]{{%
\expandafter\ifx\csname l@#1\endcsname\relax
\typeout{** WARNING: IEEEtran.bst: No hyphenation pattern has been}%
\typeout{** loaded for the language `#1'. Using the pattern for}%
\typeout{** the default language instead.}%
\else
\language=\csname l@#1\endcsname
\fi
#2}}
\providecommand{\BIBdecl}{\relax}
\BIBdecl

\bibitem{unnikrishnan2005}
R.~Unnikrishnan and M.~Hebert, ``Fast extrinsic calibration of a
  laserrangefinder to a camera,'' 07 2005.

\bibitem{scaramuzza}
D.~{Scaramuzza}, A.~{Harati}, and R.~{Siegwart}, ``Extrinsic self calibration
  of a camera and a 3d laser range finder from natural scenes,'' in \emph{2007
  IEEE/RSJ International Conference on Intelligent Robots and Systems}, Oct
  2007, pp. 4164--4169.

\bibitem{Zhang1389752}
{Qilong Zhang} and R.~{Pless}, ``Extrinsic calibration of a camera and laser
  range finder (improves camera calibration),'' in \emph{2004 IEEE/RSJ
  International Conference on Intelligent Robots and Systems (IROS) (IEEE Cat.
  No.04CH37566)}, vol.~3, Sep. 2004, pp. 2301--2306 vol.3.

\bibitem{Naroditsky2011}
O.~{Naroditsky}, A.~{Patterson}, and K.~{Daniilidis}, ``Automatic alignment of
  a camera with a line scan lidar system,'' in \emph{2011 IEEE International
  Conference on Robotics and Automation}, May 2011, pp. 3429--3434.

\bibitem{Kwak6094490}
K.~{Kwak}, D.~F. {Huber}, H.~{Badino}, and T.~{Kanade}, ``Extrinsic calibration
  of a single line scanning lidar and a camera,'' in \emph{2011 IEEE/RSJ
  International Conference on Intelligent Robots and Systems}, Sep. 2011, pp.
  3283--3289.

\bibitem{Zhou201414}
L.~{Zhou}, ``A new minimal solution for the extrinsic calibration of a 2d lidar
  and a camera using three plane-line correspondences,'' \emph{IEEE Sensors
  Journal}, vol.~14, no.~2, pp. 442--454, Feb 2014.

\bibitem{Ruben7139700}
R.~{Gomez-Ojeda}, J.~{Briales}, E.~{Fernandez-Moral}, and
  J.~{Gonzalez-Jimenez}, ``Extrinsic calibration of a 2d laser-rangefinder and
  a camera based on scene corners,'' in \emph{2015 IEEE International
  Conference on Robotics and Automation (ICRA)}, May 2015, pp. 3611--3616.

\bibitem{Huang2009}
L.~{Huang} and M.~{Barth}, ``A novel multi-planar lidar and computer vision
  calibration procedure using 2d patterns for automated navigation,'' in
  \emph{2009 IEEE Intelligent Vehicles Symposium}, June 2009, pp. 117--122.

\bibitem{PANDEY2010336}
\BIBentryALTinterwordspacing
G.~Pandey, J.~McBride, S.~Savarese, and R.~Eustice, ``Extrinsic calibration of
  a 3d laser scanner and an omnidirectional camera,'' \emph{IFAC Proceedings
  Volumes}, vol.~43, no.~16, pp. 336 -- 341, 2010, 7th IFAC Symposium on
  Intelligent Autonomous Vehicles. [Online]. Available:
  \url{http://www.sciencedirect.com/science/article/pii/S1474667016350790}
\BIBentrySTDinterwordspacing

\bibitem{Zhou201206}
L.~Zhou and Z.~Deng, ``Extrinsic calibration of a camera and a lidar based on
  decoupling the rotation from the translation,'' 06 2012, pp. 642--648.

\bibitem{Zhou201810}
L.~Zhou, Z.~Li, and M.~Kaess, ``Automatic extrinsic calibration of a camera and
  a 3d lidar using line and plane correspondences,'' 10 2018, pp. 5562--5569.

\bibitem{butvelodyne}
M.~Velas, M.~Spanel, Z.~Materna, and A.~Herout, ``Calibration of rgb camera
  with velodyne lidar.''

\bibitem{spherical2018}
J.~{Kümmerle}, T.~{Kühner}, and M.~{Lauer}, ``Automatic calibration of
  multiple cameras and depth sensors with a spherical target,'' in \emph{2018
  IEEE/RSJ International Conference on Intelligent Robots and Systems (IROS)},
  Oct 2018, pp. 1--8.

\bibitem{zhang-camera-calib}
\BIBentryALTinterwordspacing
Z.~Zhang, ``A flexible new technique for camera calibration,'' \emph{IEEE
  Trans. Pattern Anal. Mach. Intell.}, vol.~22, no.~11, p. 1330–1334, Nov.
  2000. [Online]. Available: \url{https://doi.org/10.1109/34.888718}
\BIBentrySTDinterwordspacing

\bibitem{Hartley2003MVG861369}
R.~Hartley and A.~Zisserman, \emph{Multiple View Geometry in Computer Vision},
  2nd~ed.\hskip 1em plus 0.5em minus 0.4em\relax New York, NY, USA: Cambridge
  University Press, 2003.

\bibitem{LSD4731268}
R.~{Grompone von Gioi}, J.~{Jakubowicz}, J.~{Morel}, and G.~{Randall}, ``Lsd: A
  fast line segment detector with a false detection control,'' \emph{IEEE
  Transactions on Pattern Analysis and Machine Intelligence}, vol.~32, no.~4,
  pp. 722--732, April 2010.

\bibitem{opencvlibrary}
G.~Bradski, ``{The OpenCV Library},'' \emph{Dr. Dobb's Journal of Software
  Tools}, 2000.

\bibitem{RANSAC}
\BIBentryALTinterwordspacing
M.~A. Fischler and R.~C. Bolles, ``Random sample consensus: A paradigm for
  model fitting with applications to image analysis and automated
  cartography,'' \emph{Commun. ACM}, vol.~24, no.~6, pp. 381--395, Jun. 1981.
  [Online]. Available: \url{http://doi.acm.org/10.1145/358669.358692}
\BIBentrySTDinterwordspacing

\bibitem{PCL}
R.~B. {Rusu} and S.~{Cousins}, ``3d is here: Point cloud library (pcl),'' in
  \emph{2011 IEEE International Conference on Robotics and Automation}, May
  2011, pp. 1--4.

\bibitem{Levinson201306}
J.~Levinson and S.~Thrun, ``Automatic online calibration of cameras and
  lasers,'' 06 2013.

\bibitem{ceres-solver}
S.~Agarwal, K.~Mierle, and Others, ``Ceres solver,''
  \url{http://ceres-solver.org}.

\bibitem{PribylZC16a}
\BIBentryALTinterwordspacing
B.~Pribyl, P.~Zemc{\'{\i}}k, and M.~Cad{\'{\i}}k, ``Pose estimation from line
  correspondences using direct linear transformation,'' \emph{CoRR}, vol.
  abs/1608.06891, 2016. [Online]. Available:
  \url{http://arxiv.org/abs/1608.06891}
\BIBentrySTDinterwordspacing

\bibitem{ROS}
M.~Quigley, K.~Conley, B.~Gerkey, J.~Faust, T.~Foote, J.~Leibs, R.~Wheeler, and
  A.~Ng, ``Ros: an open-source robot operating system,'' vol.~3, 01 2009.

\bibitem{Owens_msg-cal:multi-sensor}
J.~L. Owens, P.~R. Osteen, E.~Corporation, and K.~Daniilidis, ``Msg-cal:
  Multi-sensor graph-based calibration.''

\end{thebibliography}

\end{document}